# Revisiting Data Augmentation for Rotational Invariance in Convolutional Neural Networks

Facundo Quiroga[1], Franco Ronchetti[1], Laura Lanzarini[1], and Aurelio Fernandez-Bariviera[2]

[1] Instituto de Investigación en Informática LIDI, Facultad de Informática, Universidad Nacional de La Plata.
*{fquiroga,fronchetti,laural}@lidi.info.unlp.edu.ar*
[2] Universitat Rovira i Virgili, Business Management Department
*{aurelio.fernandez}@urv.cat*

**Abstract.** Convolutional Neural Networks (CNN) offer state of the art performance in various computer vision tasks. Many of those tasks require different subtypes of affine invariances (scale, rotational, translational) to image transformations. Convolutional layers are translation equivariant by design, but in their basic form lack invariances. In this work we investigate how best to include rotational invariance in a CNN for image classification. Our experiments show that networks trained with data augmentation alone can classify rotated images nearly as well as in the normal unrotated case; this increase in representational power comes only at the cost of training time. We also compare data augmentation versus two modified CNN models for achieving rotational invariance or equivariance, Spatial Transformer Networks and Group Equivariant CNNs, finding no significant accuracy increase with these specialized methods. In the case of data augmented networks, we also analyze which layers help the network to encode the rotational invariance, which is important for understanding its limitations and how to best retrain a network with data augmentation to achieve invariance to rotation.

## 1 Introduction

Convolutional Neural Networks (CNNs) currently provide state of the art results for most computer vision applications [*Dieleman et al., 2016*]. Convolutional layers learn the parameters of a set of FIR filters. Each of these filters can be seen as a weight-tied version of a traditional feedforward layer. The weight-tying is performed in such a way that the resulting operation exactly matches the convolution operation.

While feedforward networks are very expressive and can approximate any smooth function given enough parameters, a consequence of the weight tying scheme is that convolutional layers do not have this property. In particular, traditional CNNs cannot deal with objects in domains where they appear naturally rotated in arbitrary orientations, such as texture recognition *[Marcos et al. 2016]*, handshape recognition *[Quiroga et al. 2017]*, or galaxy classification *[Dieleman et al. 2015]*.

Dealing with rotations, or other set of geometric transformations, requires the network to be invariant or equivariant to those transformations. A network **f** is invariant to a transformation $\varphi$ if transforming the input to the network **x** with $\varphi$ does not change the output of the network, that is, we have **f ($\varphi$(x)) = f (x)** for all **x**. A network is equivariant to a transformation $\varphi$ if its output changes *predictably* when the input is transformed by $\varphi$. Formally, it is equivariant if there exists a smooth function $\varphi'$ such that for all **x**, we have **f ($\varphi$(x)) = $\varphi'$(f (x))** [Dieleman et al. 2016].

Depending on the application, invariance and/or equivariance may be required in different layers of the network.

While traditional CNNs are translation equivariant by design [*Dieleman et al., 2016*], they are neither invariant nor equivariant to other types of transformations in usual training/usage scenarios. There are two basic schemes for providing rotation invariance to a network: augmenting the data or the model.

Data-augmentation is a very common method for achieving invariance to geometric transformations of the input and improving generalization accuracy. Invariance and equivariance to rotations via data augmentation has been studied for Deep Restricted Boltzmann Machines *[Larochelle et al., 2007]* as well as HOGs and CNNs *[Lenc and Vedaldi, 2014]*. These results show evidence in favour of the hypothesis that traditional CNNs can learn automatically equivariant and invariant representations by applying transformations to their input. However, these networks require a bigger computational budget to train since the transformation space of the inputs must be explored by transforming them.

Other approaches modify the model or architecture instead of the input to achieve invariance. For example, some modified models employ rotation invariant filters, or pool multiple predictions made with rotated versions of the input images. However, it is unclear how and whether these modifications improve traditional CNNs trained with data augmentation, both in terms of efficiency and representational power. Furthermore, the mechanisms by which traditional CNNs achieve invariances to rotation is still poorly understood, and how as well as how best to augment data to achieve rotational invariance.

This paper compares modified CNNs models with data augmentation techniques for achieving rotational invariance for image classification. We perform experiments with various well understood datasets (MNIST, CIFAR10), and provide evidence for the fact that despite clever CNNs modifications, data augmentation is still necessary with the new models and paired with traditional CNNs can provide similar performance while remaining simpler to train and understand.

## 2 REVIEW OF CONVOLUTIONAL NEURAL NETWORKS MODELS WITH ROTATIONAL INVARIANCE

In this subsection we review the literature on modified CNN models for rotation invariance.

Many modifications for CNNs have been proposed to provide rotational invariance (or equivariance) *[Jaderberg et al., 2015, Laptev et al., 2016, Gens and Domingos, 2014, Wu et al., 2015, Marcos et al., 2016, Dieleman et al., 2016, Cohen and Welling, 2016a, Cohen and Welling, 2016b, Zhou et al., 2017]*.

Some researchers claim that for classification it is usual to prefer that the lower layers of the network encode equivariances, so that multiple representations of the input can coexist, and the upper layers of the network encode invariances, so that they can collapse those multiple representations in a useful fashion *[Lenc and Vedaldi, 2014]*. In this way, we can make the network learn all the different orientations of the object as separate entities, and then map all those representations to a single class label.

Alternatively, we can add an explicit image reorientation scheme that is applied to the image before passing it as an input to the network. In this way, the network can ignore the rotation of the object and learn a representation in a unique, canonical orientation, which simplifies the network design. However, this requires an additional model that can predict the orientation of the object.

The first approach puts the invariance near the output; the second puts it near the input by making the input layer rotation invariant. Moreover, for some objects we desire not just whole-image rotation invariance, but also invariance for some of the object's parts. For example, the arms of a person may rotate around the shoulders. It is clear that for these types of problems making the input invariant to global rotations is insufficient.

The following subsection reviews modified CNN models that deal with rotation (in,equi)-variance. We divide them into two groups: those that attempt to deal with the rotation problem globally by **transforming the input image or feature map**, and those that propose to **modify the convolution layer** in some sense to produce equivariant features and, optionally, a way to turn those features into invariant ones.

### 2.1 Transformation of the input image or feature map

*Spatial Transformer Networks (STN) [Jaderberg et al., 2015]* defines a new Spatial Transformer Layer (STL) (Figure ) that can learn to rotate input feature maps to a canonical orientation so that subsequent layers can focus on the canonical representation. Actually, STLs can also learn to correct arbitrary affine transformations by employing a sub-network that takes the feature maps as a parameter and outputs a 6-dimensional vector that encodes the affine transformation parameters. The transformation is applied via a differentiable bilinear interpolation operation. While typically the STLs are added as the first layer of the network, the layers are modular and can be added at any point in the network's convolutional pipeline.

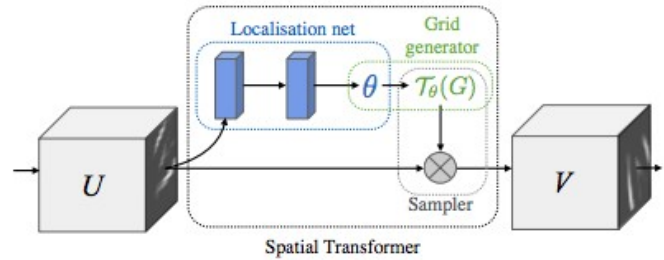

Figure : Architecture of a Spatial Transformer Layer from *[Jaderberg et al., 2015]*. The layer transform feature maps **U** to feature maps **V** by applying an affine transform **T**. The parameters of the affine transform **θ** are predicted by a localization network.

*Deep Symmetry Networks (DSN) [Gens and Domingos, 2014]* also transforms the image prior to convolution and max-pools the results, but adds an iterative optimization procedure over the 6-dimensional space of affine transformation to find a transformation that maximally activates the filter, mixing of ideas from TIP and STN. In spirit, their approach is similar to the STN approach but the optimization procedure is less elegant than the jointly trained localization network and could be seen as well as a form of data augmentation. They compare against pure data augmentation in the MNIST and NORB datasets and find that while DSNs has better performance when training with less than 10000 samples, data augmentation achieves the same performance at that point.

*Transformation-Invariant Pooling (TIP) [Laptev et al., 2016]* proposes to define a set of transformations **Φ = {φ[1] , . . . , φ[n]}** to which the network must have invariance, and then train a siamese network with a subnetwork **N[i]** for each transformation **φ[i]**. The input to the i-th subnetwork is **φ[i](x)**, that is, subnetworks share parameters but each is fed with an input transformed in a different way. A max-pooling operation is performed on the vector of outputs of the siamese network for each class. In this way the output of the network is still a vector of probabilities, one for each class. This pooling operation is crucial since it provides the invariance needed; before that operation, the representation would be (at most) equivariant.

The set of transformations **Φ** is user-defined and can include a set of fixed rotations; the authors show that whenever **Φ** forms a group then the siamese network is guaranteed to be invariant to **Φ** (assuming the input images are only affected by that set of transformations, else the network will be *approximately* invariant). TIP can be viewed as a form of test-time data augmentation, that prepares the network by data-augmentating the training of the feature extraction part of the network.

## 2.2 Modifications of the convolution layer

*Flip-Rotate-Pooling Convolutions (FRPC) [Wu et al., 2015]* extends the convolution layer by rotating the filter instead of the image. In this way, a oriented convolution generates additional feature maps by rotating traditional convolutional filters in **n*r** fixed orientations (**n*r** is a parameter). Then, max-pooling along the channel dimension is applied to the responses of the **n*r** orientations so that a single feature map results. No comparison to data augmentation approaches was performed. While the number of parameters of the layer is not increased, the runtime memory and computational requirements of the layer are multiplied by **n*r**, although the number of parameters for the full network can actually be reduced since the filters are more expressive.

The same approach is used in *[Marcos et al., 2016]* for texture classification; they do perform comparisons with data augmentation but only for 20 samples. Given that the rotation prior is very strong in texture datasets, this is an unfair comparison.

*Exploiting Cyclic Symmetry in CNNs [Dieleman et al., 2016]* presents a method similar to FPRC alongside variants that provide equivariance as well as invariance. *Oriented response networks (ORN) [Zhou et al., 2017]* are related to FRPC, but also introduce an ORAlign layer that instead of pooling the set of features maps generated by a rotating filter reorders them in a SIFT-inspired fashion which also provides invariance.

*Dynamic Routing Between Capsules [Sabour et al., 2017]* presents a model are units of neurons designed to mimic the action of cortical columns. Capsules are designed to be invariant to complicated transformations of the input. Their outputs are merged at the deepest layer, and so are only invariant to global transformation

*Group Equivariant Convolutional Networks* (GCNN) and *Steerable CNNs [Cohen and Welling, 2016a,Cohen and Welling, 2016b]* also use the same basic methods but provide a more formal theory for guaranteeing the equivariance of the intermediate representations by viewing the set of transformations of the filters as a group. *Learning Steerable Filters for Rotation Equivariant CNNs [Weiler et al, 2018]* also employs the same approach. *Spherical CNNs [Cohen et al, 2018]* extend this approach to the 3D case.

In particular, Group CNNs *[Cohen and Welling, 2016a]* adds an additional *rotation* dimension to the convolutional layer. This dimension allows to compute rotated versions of the feature maps in 0°, 90°, 180° and 270° orientations, as well as their corresponding horizontally flipped versions. The first convolution *lifts* the image channels by adding this dimension; further convolutions compute the convolution across all channels and rotations, so that the filter parameters are of size **Channels\*Rotations\*H\*W**, where H and W are the height and width of the filter. The bias term is the same for all rotation dimensions. To *return* to the normal representation of feature maps, the rotation dimension is reduced via a max operation, obtaining invariance to rotation.

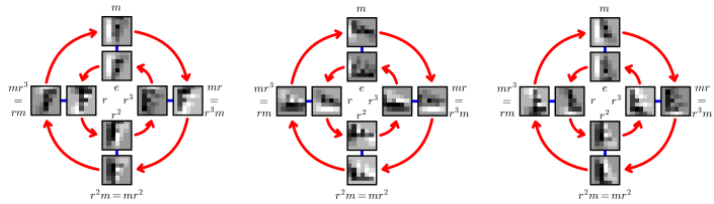

Figure . Transformations of a filter made by a Group CNN from *[Cohen and Welling, 2016a]*. Filters are rotated and flipped and then applied to the feature maps. The rotation and flip operations form an algebraic group.

We note that all previous models are variations of the same strategy: augment convolutions with equivariance to a finite set of transformations by rotating or adapting the filters, then provide a method to collapse the equivariance into an invariance before the final classification.

*Deformable Convolutional Networks [Dai et al., 2017]* learns filters of arbitrary shape. Each position of the filter can be arbitrarily spatially translated, via a mapping that is learned. Similarly to STN, the employ a differentiable bilinear filter strategy to sample from the input feature map. While not restricted to rotation, this approach is more general than even STN since the transformation to be learned is not limited by any affine or other priors; hence we do not consider it for the experiments.

## 3 EXPERIMENTS

We performed two types of experiments understand data augmentation for rotational invariance and compare it with other methods. We used the MNIST and CIFAR10 datasets (Figure 1) in our experiments [Le Cunn et al. 1998, Alex Krizhevsky et al. 2009] because these are well known and the behavior of common networks such as those tested here is better understood than for other datasets, and cover both synthetic grayscale images and RGB natural images.
.

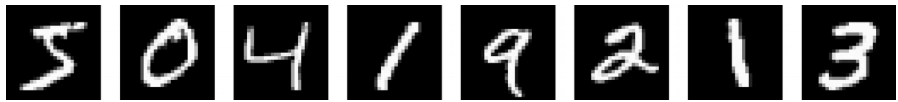

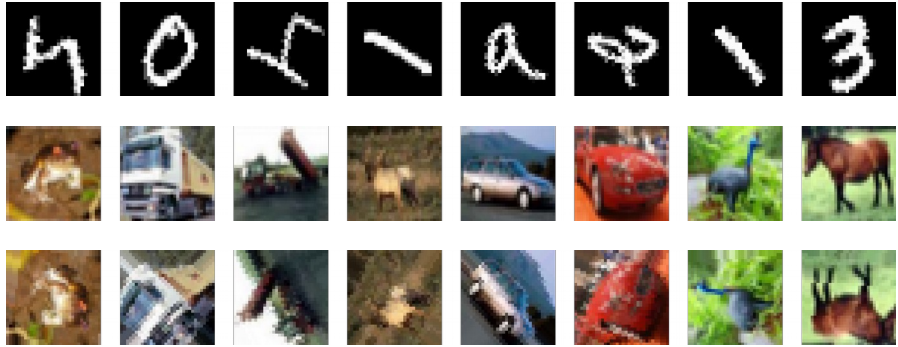

Figure 1: **Rows** 1,2) MNIST and rotated MNIST images. 3,4) CIFAR10 and rotated CIFAR10 images. Note that for CIFAR10 some images are slightly cropped by the rotation procedure, while for MNIST the cropping is negligible due to the black border.

The data augmentation we employed consists of rotating the input image by a random angle in the range [0°,360°]. For the MNIST dataset, previous works employ the MNISTrot version in which images are rotated in 8 fixed angles. While using a discrete set of rotations allows some models to learn a fixed set of filters for guaranteed equivariance, we chose to use a continuous set of rotations because it better reflects real-world usage of the methods. We tested only global rotations; ie, rotations of the whole image around the center.

In all experiments networks were trained until convergence by monitoring the test accuracy, using the ADAM optimization algorithm with a learning rate of 0.0001 and 1E-9 weight decay for all non-bias terms.

**3.1 NETWORK MODELS**

We employed a simple convolutional network we will call **SimpleConv**, that while it clearly does not provide state of the art results, is easy to understand and well . The simple convolutional network for is defined as: Conv1(F) - Conv2(F) – MaxPool1(2x2) – Conv3(F*2) – Conv4(F*2) – MaxPool2(2x2) – Conv5(F*4) – FC1(D) – Relu – BatchNorm – FC(10), where all convolutional filters are 3x3, and there is a ReLU activation function after each convolutional and fully connected . F is the number of feature maps of the convolutional layers and D the hidden neurons of the fully connected layer. For MNIST, we set F=32 and D=64, while for CIFAR10 F=64 and D=128, matching the original Group CNN implementation (see below).

To test the importance of the Dense layers in providing invariance to rotated samples, we also experimented with an **AllConvolutional** network. This model uses just convolutions and pooling as building blocks, which is of interest to our .

The AllConv network is defined as: Conv1(F) – Conv2(F) – Conv3(F,stride=2x2) - Conv4(F*2) – Conv5(F*2) – Conv6(F*2,stride=2x2) – Conv7(F*2) – Conv8(F*2, 1x1) - ConvClass(10, 1x1) – GlobalAveragePooling(10). Again, all convolutional filters are 3x3, and we place ReLUs after convolutions. For MNIST, F=16 while for CIFAR10, F=96, again matching the original Group CNN implementation.

Then, we chose a model from the two groups described in section 2 - transformation of the input image and transformation of the filters – that also correspond to the alternative strategies of putting the invariance near the input or near the output.

As a representative of the first group we added a **Spatial Transformer Layer** *[Jaderberg et al., 2015]* to the convolutional network to reorient the image before the network classifies it; the resulting networks are named **SimpleConvSTN** and **AllConvolutional STN**. To keep comparisons fair, we modified the localization and affine matrix to restrict transformations to rotations. The localization network consists of a simple CNN with layers: Conv1(16,7x7) – MaxPool1(2x2) – ReLU() - Conv2(16,5x5) - MaxPool2(2x2) – ReLU() - FC(32)

From the modified convolutional methods, we chose **Group CNNs** *[Cohen and Welling, 2016a]* . The resulting networks are named **SimpleGConv** and **AllGConvolution**, where we simply replaced normal convolutions for group convolutions, and we added a pooling operation before the classification layers to provide the required invariance to rotation. Group CNNs has 4 times the feature maps as a regular CNN network; to compensate, and as a compromise, we reduced the parameters by half.

### 3.1 DATA AUGMENTATION WITH TRADITIONAL NETWORKS

First we measured the performance of a SimpleConv and AllConvolutional with and without data augmentation to obtain a baseline for both base methods.

We trained two instances of each model; one with the normal dataset, and the other with a data-augmented version. We then tested each instance of each model with the test set of the normal and data-augmented variant. Figure 2 shows the results of the experiments for each model/dataset combination.

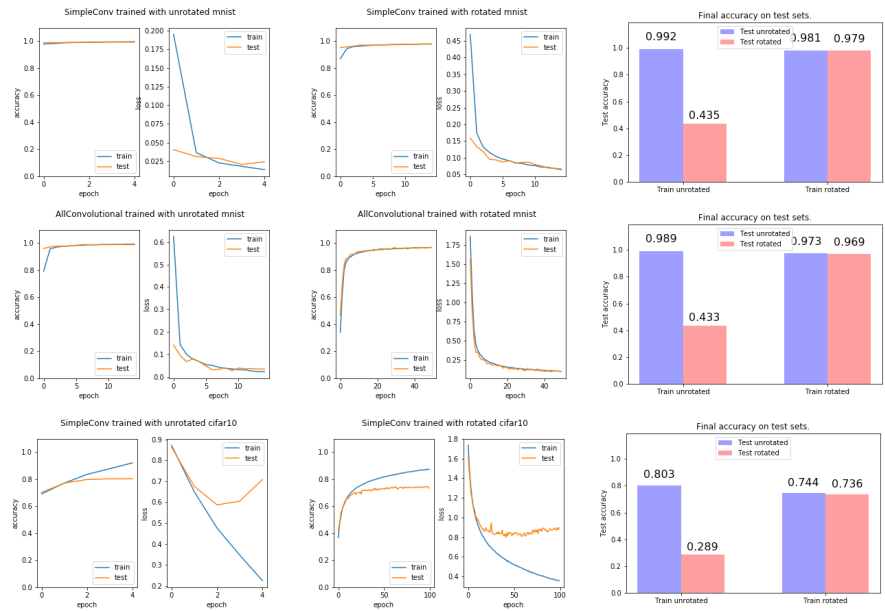

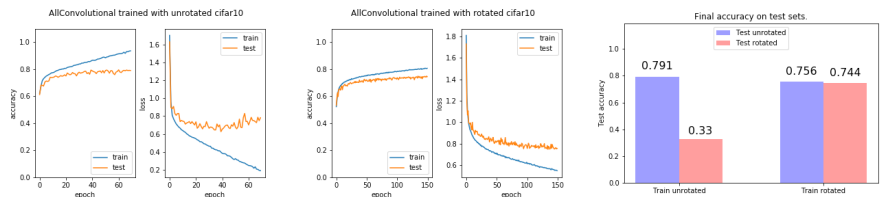

Figure 2: **Rows** 1,2) SimpleConv and AllConvolutional with MNIST. 3,4) SimpleConv and AllConvolutional with CIFAR10. **Left** Accuracy and loss for each training epoch, on training and test set. **Middle** Same as left, but with a rotated training set. **Right** Final test set accuracies for the two models trained and the two variations of the dataset (unrotated, rotated).

On MNIST, we can observe that while networks trained with a rotated dataset see a drop in their accuracies of 1%-2% on the unrotated test set, the performance drop for networks trained with the unrotated dataset and tested with the rotated dataset fare much worse (~55% drop in accuracy). It is surprising that the drop in the first case is so low, specially given that the number of parameters is the same for both networks. This may indicate a redundancy in the filters of the unrotated model. Also, it would seem that the network trained on the unrotated dataset still performs at a ~40% accuracy level, four times more than expected by chance (10%). This is partly because some of the samples are naturally invariant to rotations such as the images for the number 0 or invariant to some rotations such as the numbers 1 or 8, and because some of the learned features must be naturally invariant to rotation as well.

In the case of CIFAR10 the results show a similar situation, although the drop in accuracy from unrotated to rotated is larger, possibly because the dataset possesses less natural invariances than MNIST. Note that to reduce the burden of computations the number of training epochs on CIFAR10 with AllConvolutional was reduced, achieving ~80% accuracy instead of 91% as in the authors original experiments *[Jaderberg et al., 2015]*.

Still, it is surprising that the AllConvolutional networks can learn the rotated MNIST dataset so well, since convolutions are neither invariant nor equivariant to rotation. This points to the fact that the set of filters learned by the network possibly self-organize during learning to obtain a set of filters that can represent all the rotated variations of the object.

### 3.2 COMPARISON WITH STN AND GROUP CNN

We ran the same experiment as before but using the modified versions of the network. Figure 3 shows the results of the STLs versions of the networks.

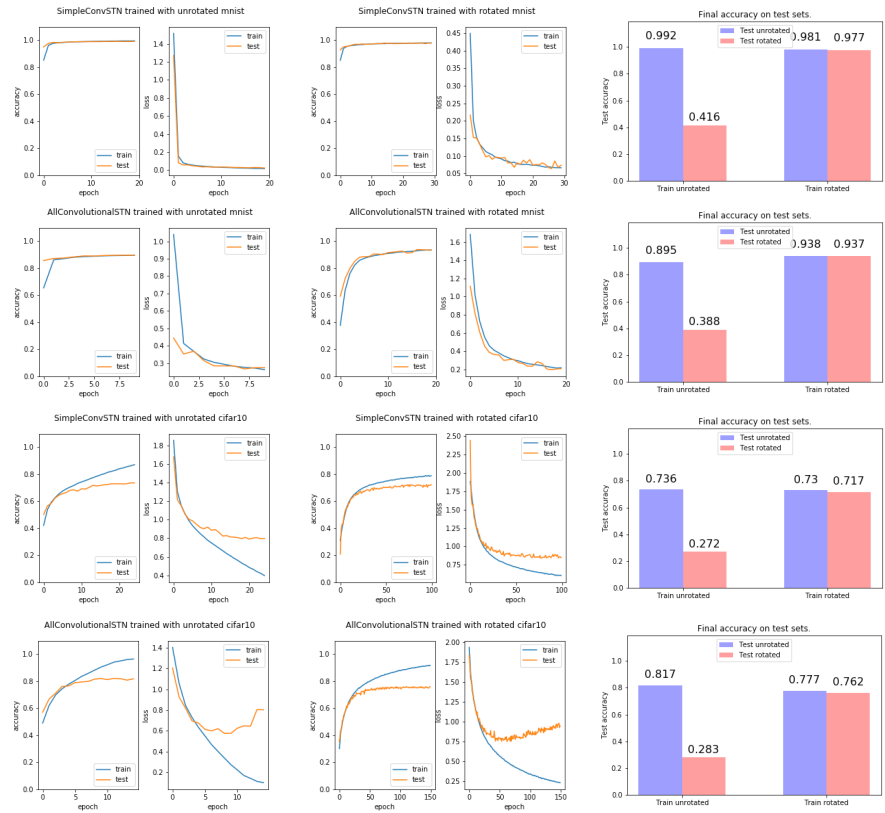

Figure 3: **Rows** 1,2) SimpleConvSTN and AllConvolutionalSTN with MNIST. 3,4) SimpleConvSTN and AllConvolutionalSTN with CIFAR10. **Left** Accuracy and loss for each training epoch, on training and test set. **Middle** Same as left, but with a rotated training set. **Right** Final test set accuracies for the two models trained and the two variations of the dataset (unrotated, rotated)

We can see that in all cases the performance of the unrotated model on the rotated dataset is much lower than the original; this is expected since the STL needs data augmentation during training. However, the STL model did not perform noticeable better than the normal data-augmentated models (Figure 2).

Figure 4 shows as well the results on MNIST and CIFAR10 of the Group CNN models. Similarly to the STN case, the performance is not increased with Group CNN: however, in the case of the AllGConvolutional network we see that the performance of the model trained with the unrotated dataset on the rotated dataset is much greater (+0.1) than for other models, while the same does not happen with the SimpleGConv network. This is possibly due to the fact that the superior representation capacity of the fully convolutional layers can compensate for absence of good filters, while in the AllGConvolutional case there is more pressure on the convolutional layers to learn good representations.

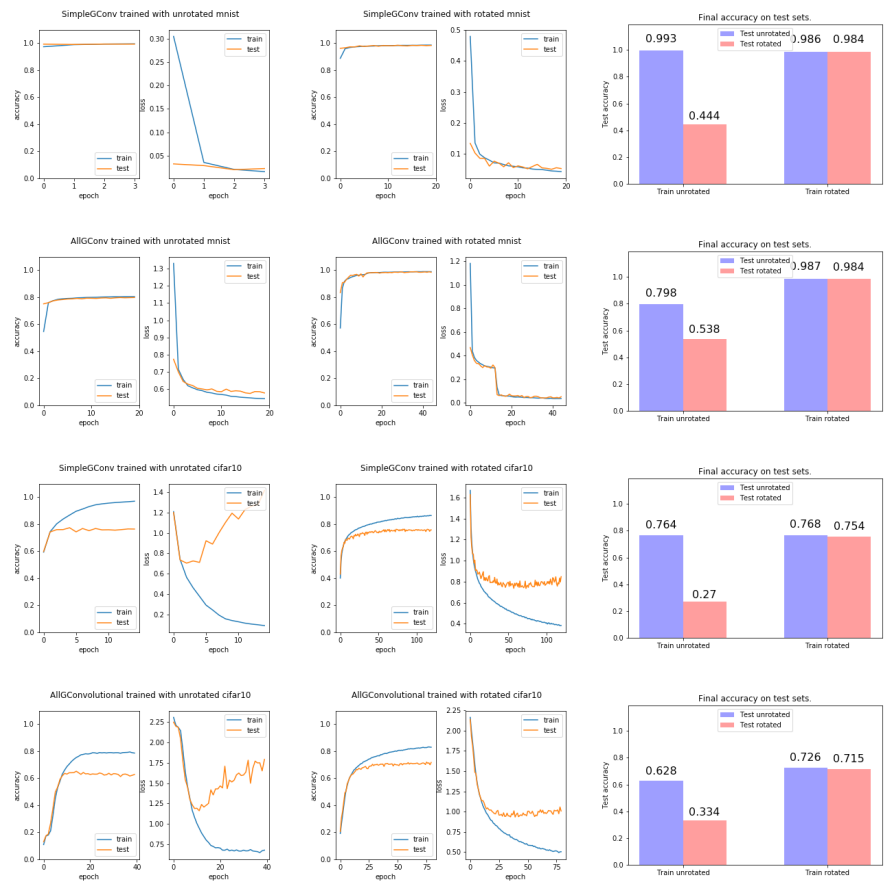

Figure 4: **Rows** 1,2) SimpleGConv and AllGConvolutional with MNIST. 3,4) SimpleGConv and AllGConvolutional with CIFAR10. **Left** Accuracy and loss for each training epoch, on training and test set. **Middle** Same as left, but with a rotated training set. **Right** Final test set accuracies for the two models trained and the two variations of the dataset (unrotated, rotated)

### 3.3 RETRAINING FOR ROTATIONAL INVARIANCE

While section 3.2 seems to point to the fact that there is small difference in accuracy when using specialized models versus data augmentation, it could be argued that specialized models can be more efficient when training. Alternatively, if training time is a limiting factor, we could use a pretrained network and retrain some of its layers to achieve invariance. However, a priori it is not clear whether the whole network can/needs to be retrained, or if only some parts of it need to adapt to the rotated examples.

To analyze which layers are amenable to retraining for rotation invariance purposes, we train a base model with an unrotated dataset, then make a copy of the network and retrain parts of it (or all of it) to assess which layers can be retrained.

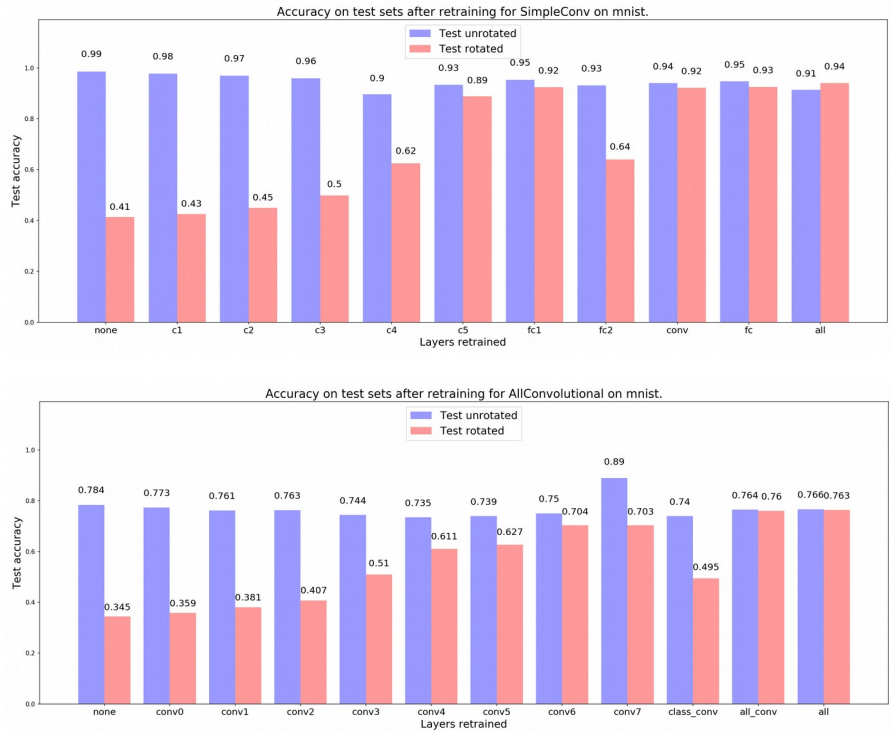

Figure 5: Retraining experiments for SimpleConv and AllConvolutional on MNIST. Accuracy on the test sets after retraining subsets of layers of the network. The labels "conf"/"all_conv" and "fc" mean that all convolutional or fully connected layers were retrained.

As Figure 5 shows, retraining for rotation invariance shows a similar trend to retraining for transfer learning; higher layers with more high-level features are a better target for retraining individually since they are closer to the output layers and can affect them more. The fact that retraining the penultimate layers can bring performance back means that there's a redundancy of information in previous layers since rotated versions of the images can be reconstructed from the output of the first layers. Since the original networks were trained with unrotated examples, this means that either that network naturally learns equivariant filters, or that equivariance in filters is not so important for classifying rotated objects.

However, it is surprising that retraining the final layer in both cases leads to a reversal of this situations: the performance obtained is less than when retraining other layers. In the case of the SimpleConv network, this is possibly due to the action of previous fc layer collapsing the equivariances before the final layer can translate them to a decision; that is, the fc1 layers must be loosing *some* information. In the case of the AllConvolutional network, the class_conv layer performs a simple 1x1 convolution to collapse all feature maps to 10, and so probably cannot recapture the invariances, while the retraining the previous layer with many more 3x3 convolutions can.

Figure 6 shows the results of the same experiment on CIFAR10. The results are similar to those of MNIST, except for the lower general accuracy given the difficulty of CIFAR10.

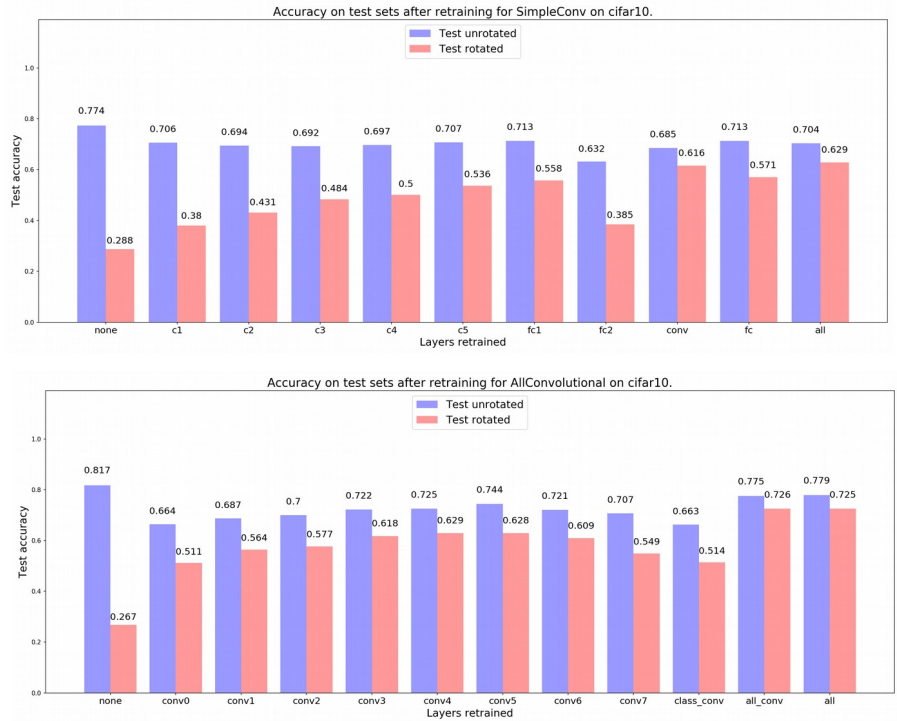

Figure 5: Retraining experiments for SimpleConv and AllConvolutional on CIFAR10. Accuracy on the test sets after retraining subsets of layers of the network. The labels "conf"/"all_conv" and "fc" mean that all convolutional or fully connected layers were retrained.

## 4 CONCLUSIONS

Rotational invariance is a desired property for many applications in image classification. Data augmentation is a simple way of training CNNs models, which currently hold the state of the art, to achieve invariance. There are many modified CNNs models that attempt to make this task easier.

We compared data augmentation with modified models, maintaining the same number of parameters in each case. While data augmentation requires more training, it can reach similar accuracies as other methods. Furthermore, the test time is not affected by additional localization networks or convolutions, and

We also performed retraining experiments with data augmentation to shed some light on how and where can network learn rotational invariance. By retraining layers separately, we found that some invariance can be achieved added in every layer, although the ones nearest the end of the network, whether fully connected or convolutional, are much better at gaining invariance. This finding reinforces the notion that lower layers of networks learn redundant filters and so can be capitalized for other tasks (rotation invariant tasks in this case) and also the notion that invariance should be added at the end of the network if possible.

We believe more can be learned about CNNs by studying their learned invariances. Possible extensions of this work include deliberately introducing invariances early or late in the network to see how they affect the capacity of the network to learn rotations. It would be useful as well to compare the convolutional filters learned to identify equivariances between their outputs, as well as see via their activations which are rotationally invariant. Systematic experimentation on datasets where samples are rotated naturally, as well as naturally rotation invariant, and comparison to datasets where the rotation is synthetic such as those in this work are needed. To reduce the experimentation burden, we have chosen to keep the number of parameters constant in the experiments, but it would also be desirable to see the impact of data augmentation when the number of parameters is constrained.